%% file: main.tex
\documentclass[sigconf]{acmart}
\renewcommand\footnotetextcopyrightpermission[1]{} % removes footnote with conference information in first column
\usepackage{pifont} % used for \todo symbol

\usepackage{hyperref}
\usepackage{float}
\usepackage{multirow}
\usepackage{multicol}
\usepackage{array}

\def\BibTeX{{\rm B\kern-.05em{\sc i\kern-.025em b}\kern-.08emT\kern-.1667em\lower.7ex\hbox{E}\kern-.125emX}}
    
\begin{document}

\title{Collaborative Machine Learning-Driven Internet of Medical Things - A Systematic Literature Review}

\author{Rohit Shaw}
\email{rohit.shaw@student.uva.nl}
\affiliation{
  \institution{Universiteit van Amsterdam}
}

\begin{abstract}
The growing adoption of IoT devices for healthcare has enabled researchers to build intelligence using all the data produced by these devices. Monitoring and diagnosing health have been the two most common scenarios where such devices have proven beneficial. Achieving high prediction accuracy was a top priority initially, but the focus has slowly shifted to efficiency and higher throughput, and processing the data from these devices in a distributed manner has proven to help achieve both. Since the field of machine learning is vast with numerous state-of-the-art algorithms in play, it has been a challenge to identify the algorithms that perform best in different scenarios. In this literature review, we explored the distributed machine learning algorithms tested by the authors of the selected studies and identified the ones that achieved the best prediction accuracy in each healthcare scenario. While no algorithm performed consistently, Random Forest performed the best in a few studies. This could serve as a good starting point for future studies on collaborative machine learning on IoMT data.

\end{abstract}


\maketitle
\pagestyle{plain}

\input{introduction.tex}
\input{study_design.tex}
\input{results.tex}
\input{discussion.tex}
\input{conclusion.tex}

\bibliographystyle{ACM-Reference-Format}
\bibliography{bibliography}

\end{document}

%% file: introduction.tex
\section{Introduction}
Over the last few years, we have seen significant growth in the number of Internet of Things devices used in the healthcare industry. The widespread adoption of self-tracking smart wearable devices such as smartwatches has been a major contributor. Since such devices produce massive amounts of data, researchers and industry specialists have been exploring ways to make efficient use of them by building (artificial) intelligence into these devices via means of machine learning and making smart devices smarter. Given the small form factor of these devices, they are often limited in terms of hardware capabilities to carry out machine learning tasks on-device. Researchers have been exploring ways to tackle this problem through distributed machine learning techniques.

This systematic study reviews the trends of adoption of IoT systems in healthcare that utilize distributed machine learning and the state-of-the-art algorithms and frameworks used by those systems. We then identify the algorithms that perform best in terms of prediction accuracy in different healthcare scenarios in the subsequent sections. The study ends with a discussion of the findings and where this area of research could potentially be heading.

%% file: study_design.tex
\section{Study Design}\label{sec:design}
This study was carried out by setting a research goal and determining research questions to help achieve it. The query string was then run on Google Scholar to obtain preliminary results on the studies available on the topic. To narrow down the scope of the search, inclusion and exclusion criteria were determined based on the research goal and questions, which helped remove studies that were irrelevant and select primary studies. The primary studies were then deeply analyzed and relevant data from them was extracted to help answer the research questions framed.

\label{sec:design}

\subsection{Research Goal}
This research was designed to create a path for researchers and practitioners to build upon and potentially introduce new technical contributions based on the results. We identify and compare popular distributed machine learning algorithms and frameworks used in state-of-the-art IoT healthcare systems based on their performance in terms of prediction accuracy. Since there is no one-solution-fits-all framework or algorithm available, we evaluate the performance of each in different scenarios and identify those that produce the best prediction results for each healthcare scenario.

\subsection{Research Questions}
To clarify the direction of this literature review, research questions were determined along with their rationales. The research questions drove the whole study and had a special influence on 
\begin{enumerate}
    \item the search and selection of primary studies,
    \item the data extraction process,
    \item data synthesis and analysis.
\end{enumerate}

\label{sec:questions}
\begin{enumerate}
    \item[\textbf{{RQ}1:}] \textit{How has the healthcare industry been combining Machine Learning and the Internet of Things in recent years?}
    \item[{ }] The number of connected IoT devices generating health-related sensor data has grown exponentially in recent years. To make effective and efficient use of the massive amounts of data produced by these devices, various forms of AI - machine learning techniques, in particular, are being used to make smart personalized predictions. To get better insights, we review some of the state-of-the-art solutions being used in real-world applications and their evolution trends in the healthcare industry in the period 2016-2020.
\end{enumerate}

\begin{enumerate}
    \item[\textbf{{RQ}2:}] \textit{Can Distributed Machine Learning be applied to IoT Healthcare data?}
    \item[{ }] The most popular machine learning algorithms are only capable of handling small to medium-sized datasets. As the amount of data generated by IoT healthcare devices is bound to grow in volume over time, processing this data will become more and more computationally challenging. To continue driving quality predictions, all of the data must be leveraged. To mine data efficiently, the machine learning workload will have to be distributed across multiple processing agents.
\end{enumerate}

\begin{enumerate}
    \item[\textbf{{RQ}3:}] \textit{Which Distributed Machine Learning algorithms are the most suitable for processing IoT data in each healthcare scenario?}
    \item[{ }] Since no machine learning algorithm performs consistently and fits all scenarios in healthcare, we compare algorithms and distributed learning frameworks most used in state-of-the-art IoT healthcare systems based on their performance and the quality of the predictions they produce. The identification of such works and cutting-edge techniques would create a path for future technical contributions that build upon and improve existing solutions.
\end{enumerate}

\subsection{Initial Search}
This stage aimed to perform an automatic search on an electronic database and retrieve potential publications touching upon the topic of this study. Google Scholar was chosen for this as it searches across several academic and scientific databases. Google Scholar has gained credibility over the years as it been observed to have coverage advantages over controlled databases \cite{GoogleScholar}, making it a good choice for this time-boxed literature review. Next, keywords and query strings were identified and extracted from the research questions. The query string helped narrow down the scope and area of the search. The following query string was used for finding literature for this study. \\

\noindent \texttt{(Machine Learning) \textbf{AND} (Internet of Things \textbf{OR} IoT \textbf{OR} IoMT) \textbf{AND} (Health \textbf{OR} Healthcare \textbf{OR} Medicine \textbf{OR} Medical) \textbf{AND} (Distributed \textbf{OR} Federated) \textbf{AND} (Algorithm)} \\

After running the aforementioned query string and obtaining 115 results, publications that were outside the scope of this literature study were filtered out. Results such as articles, books, workshop proceedings, and the likes were removed and only 61 research papers were retained. From the remaining results, duplicates were then identified and removed, leaving us with 51 studies to apply the selection criteria to, discussed in the next section.

\begin{figure}[h!]
    \centering
    \includegraphics[width=8.7cm]{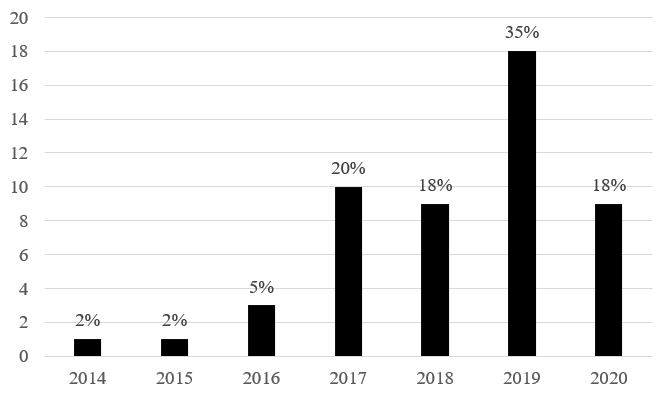}
    \caption{Distribution of Selected Studies Over Years}
    \label{fig:years_all}
\end{figure}

\subsection{Application of Selection Criteria}
As mentioned in \cite{SysReviewProc}, the selection criteria are defined when defining the execution protocol of the systematic review. The 11 quality assessment criteria methodology by \cite{ApplyingSystematicReviews} was used to define the inclusion \textit{(I)} and exclusion \textit{(E)} criteria to enable us to narrow the scope further down and only select studies that were relevant.

\begin{enumerate}
\item[\textbf{{I}1:}] Studies that address machine learning in the field of IoT.
\item[\textbf{{I}2:}] Studies that address IoT-based healthcare systems.
\item[\textbf{{I}3:}] Studies that propose a new framework, system, architecture, tool, algorithm, or methodology involving collaborative machine learning in healthcare.
\item[\textbf{{I}4:}] Studies published in the domain of artificial intelligence and healthcare.
\item[\textbf{{I}5:}] Studies developed by either academics or industry.
\item[\textbf{{I}6:}] Studies written in English. 
\item[\textbf{{E}1:}] Studies that do not consider the use of machine learning with IoT device data.
\item[\textbf{{E}2:}] Studies mainly targeting the hardware architectural side of IoT in healthcare or distributed machine learning. 
\item[\textbf{{E}3:}] Studies not available as full-text.
\end{enumerate}

After filtering out selected papers using the defined inclusion and exclusion criteria, we were left with 28 publications. However, in order to carry out a valid systematic literature review, according to \cite{ApplyingSystematicReviews}, 15-20 publications should be selected as primary studies. Looking into and assessing the abstract, introduction, results, and conclusion of each, 21 publications were selected as primary studies for this review.

\begin{figure}[h!]
    \centering
    \includegraphics[width=8.7cm]{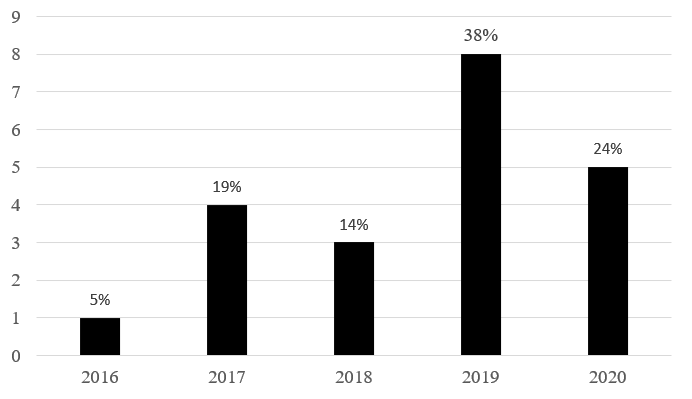}
    \caption{Distribution of Primary Studies Over Years}
    \label{fig:years}
\end{figure}

\subsection{Data Extraction and Synthesis}
In this stage, the data from the primary studies were prepared for in-depth analysis to help answer the research questions. The type of publication was noted, and the studies were summarized based on the IoMT scenarios covered, distributed computing frameworks proposed, and the distributed machine learning algorithms used.

\begin{figure}[h!]
    \centering
    \includegraphics[width=8.7cm]{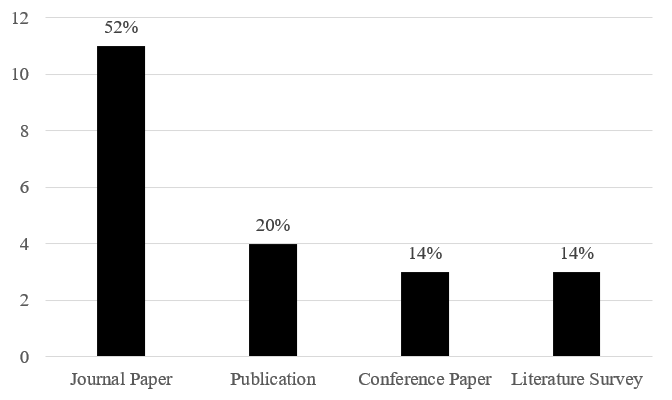}
    \caption{Distribution of Primary Study Types}
    \label{fig:study_type}
\end{figure}

Based on the results of the data extraction process, each research question was answered extensively, limitations were identified, and conclusions were derived.

%% file: results.tex
\section{Results}
In this section, the findings and results of the systematic literature review are reported. To make it easy to follow, this section has been split into three, each answering a research question.

\subsection{ML \& IoT in the Healthcare Industry Today}
\begin{figure}[h!]
    \centering
    \includegraphics[width=8.7cm]{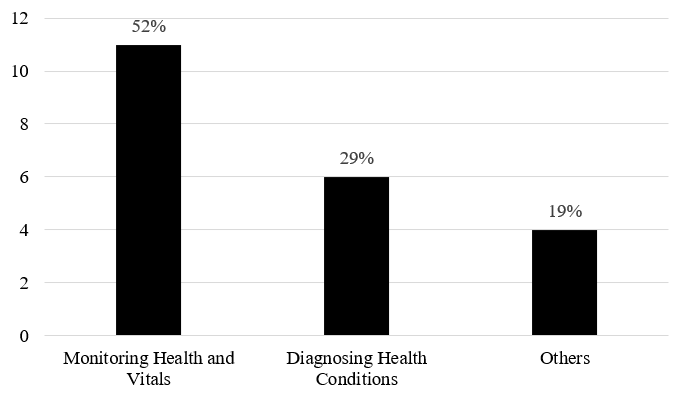}
    \caption{Number of Studies in Each Category}
    \label{fig:categories}
\end{figure}

After analyzing the primary studies for the ways in which ML and IoT are being combined in the healthcare industry today, a pattern was observed. The selected studies were then further classified based on categories created for each pattern as depicted in Figure \ref{fig:categories}.

\subsubsection{\textbf{Monitoring Health and Vitals}}
The most popular application combining ML and IoT in healthcare was monitoring the health condition and vitals of patients and quantified selves. IoT technology, although fairly new, has increasingly been gaining popularity due to its potential in a world with a rising number of people with chronic illnesses. This along with the aging population has been identified to be placing significant stress on the healthcare system due to rising resource demands from them. The standard model for IoT healthcare systems proposed by \cite{Seven} suggested a rehabilitation system for people suffering from knee injuries which would enable the calculation of the position and angle of knees, and assist the patient with knee-related motor movements, all while monitoring the knee position and angle. The system could also be adapted for various other monitoring purposes including but not limited to the management of hypertension and monitoring critical health metrics such as pulse, blood pressure, body temperature, etc \cite{Seven}\cite{Eight}. Similarly, several parameters such as physiological parameters, speech, posture, skin, motion, and environmental conditions were monitored in \cite{ThirtySix} using various machine learning techniques. \par

According to the study performed by \cite{Sixteen}, poor medication adherence was identified as another leading factor significantly stressing the healthcare system. Medication adherence was measured on how closely patients followed their prescribed medication regimens including the time and dosage. To combat this by capitalizing on IoT and ML, \cite{Sixteen} developed a distributed cloud-based medication intake monitoring system paired with an Android smartwatch application that collected sensor data. The holistic multi-layer architecture presented by \cite{Thirteen} took things a step further by enabling real-time insights, helping improve the decision-making abilities of patients as well as healthcare providers. It not only employs machine learning on the cloud but also utilizes an artificial neural network for performing ML on-chip, all while ensuring less computation intensity to compensate for the limitations of low-end battery-powered devices. \par

In the typical cloud, fog, edge computing IoT stack, fully distributed fog-based systems were observed to offer unbound and limit-free operations but at the cost of accuracy due to computation limitations at the edge layer. To combine the best of both fog and cloud, a novel hierarchical fog-assisted computing architecture (HiCH) was presented by \cite{TwentyEight}, aimed at hierarchical partitioning of ML algorithms across all layers. A linear Support Vector Machine (SVM) classifier was chosen given its low computation cost and minimal performance and accuracy loss, to detect arrhythmia. The SVM classifier returned good results, but they certainly were not at the level and quality offered by deep learning algorithms. Since IoT-based health systems offer the ability to monitor patients beyond general clinical settings, results of the highest quality are desired as even a minor flaw in the results could prove fatal for patients. The authors of \cite{ThirtyTwo} presented a system that capitalized on the HiCH architecture by \cite{TwentyEight} and deployed a Convolutional Neural Network (CNN) classifier to enable deep learning. The proposed system was used to classify ECG signals to better monitor and detect arrhythmia. The federated learning framework proposed by \cite{Twenty} follows a similar approach but without HiCH. It partitions and allocates deep neural networks to IoT devices and a centralized server while significantly minimizing computation load on both. However, this approach resulted in a small accuracy loss when deployed on a real-world arrhythmia detection system. Comparing the three systems proposed above, the one proposed by \cite{TwentyEight} was observed to offer the best mix of both performances as well as accuracy. \par

Health monitoring solutions were not only limited to patients and quantified selves. Studies \cite{Six} and \cite{Eighteen} took things to the battlefield to track the location and monitor the health of soldiers who got lost, injured on the battlefield, or are deployed in adverse environmental conditions. This was achieved by the use of Wireless Body Area Networks (WBAN) comprising various sensors. The systems not only monitored the health metrics of soldiers, but also tracked their position, helped detect nearby bombs, and predicted potential war zones, helping army control units minimize operation times and plan their actions quicker \cite{Six}. \par

The real-time medical emergency response system proposed by \cite{Fifty} also used a WBAN comprising of a plethora of sensors, data generated by which was analyzed, and necessary actions were taken accordingly. The proposed system provides direct connectivity to the end-user to avail various kinds of medical help such as hospitals and ambulances, and also police stations. The connected entities are automatically contacted in the event of an emergency such as a car accident.

\subsubsection{\textbf{Diagnosing Health Conditions}}
The second most common application observed was the health diagnosis of patients as well as predicting health conditions based on the results of the diagnosis or otherwise. Not to intertwine with health monitoring which is mainly aimed at providing real-time insights, health diagnosis takes into account a plethora of data points to analyze and comprehensively assess the health of an individual and predict potential diseases. Early detection of diseases would lead to quicker actions being taken to treat them. The cloud-centric IoT-based m-healthcare disease diagnosing framework proposed by \cite{Nine} does exactly this as well as identifying the level of the severity, enabling healthcare providers to provide the correct treatment. A smart student healthcare prototype was designed to test this framework, which employed a Decision Tree (DT), k-Nearest Neighbor (k-NN), Naïve Bayes (NB), and SVM for disease classification along with the corresponding levels of severity. If the results of a diagnosis indicated an emergency, the nearest hospital was immediately notified to handle the emergency \cite{Nine}. In the journal paper by \cite{Eleven}, a similar but scalable cloud-based teleophthalmology architecture was proposed for predicting age-related macular degeneration (AMD) disease. The retinal images of patients were diagnosed by a modified ResNet CNN (AMD-ResNet) in the cloud to identify the level of AMD severity with high precision. Precision is critical in this scenario as the medication dosage would have to be prescribed accordingly. \par

The ML algorithms employed in the works mentioned above provided good results and predictions, but as discussed in Section 3.1.1, deep learning networks were identified to offer the highest quality and the most accurate predictions. The HealthFog framework by \cite{ThirtyFive} combines the power of complex deep learning networks with cloud-based edge computing devices to detect heart-related diseases with high accuracy and very low latency. The framework was built for efficient heart-related patient data management and diagnosing heart health while also identifying disease severity. \par

The rise in the number of wearable pieces of technology has made it easy for people to get access to their health data of all sorts. The large amount of sensor data generated by these wearables could be capitalized effectively by feeding it to ML models. However, the problem of data islands and data security and privacy come into play when doing this. FedHealth, a federated transfer learning framework was designed by \cite{FiftyOne} to overcome these challenges. By aggregating the sensor data via means of federated learning and building custom models by transfer learning, it provided quality results without hampering security and privacy. Given how general this framework is, it could easily be built upon and applied to various healthcare applications and scenarios. \par

Another approach to processing healthcare data is via means of big data technologies. These can handle extensive amounts of data efficiently. The works by \cite{FortyOne} and \cite{FortyFour} make use of various big data technologies such as Spark and Kafka to apply distributed machine learning to healthcare data to perform diagnoses and drive predictions. While these technologies ace when it comes to processing large amounts of data, they are often resource-hungry and need powerful computing hardware. These will be discussed in-depth in the coming sections.

\subsubsection{\textbf{Other Innovative Systems}}
Finally, a handful of new and unique applications and systems were also identified. The majority of the solutions discussed in the previous sections were built on the cloud, fog, edge IoT computing stack and employed machine learning on it. The architectures and frameworks proposed were tailored towards handling one kind of a problem - diagnosing a particular type of disease, for instance. While these contributions were breakthroughs in their specific niches, a more complex system comprising various IoT components was not something these were aimed at or thought of. This is where \cite{TwentyNine} studied and developed a health-related data sharing system between IoT devices and a central system, using the distributed ledger technology (DLT). The system benefitted from DLT by making it secure and highly scalable, two traits that are highly desired in a data-sharing system. This architecture could enable more comprehensive IoT systems comprising of various sensors to be designed for all kinds of diagnoses, and eventually, a single system that could diagnose all possible health conditions. \par

While the concept is a single system for everything sounds both interesting and complex, \cite{TwentyFive} built on this idea and presented a holistic architecture for an IoT eHealth ecosystem based on the cloud, fog, edge computing stack. Such an ecosystem would comprise various connected IoT-based medical devices monitoring a patient at the hospital for instance. This architecture would also enable a seamless transition from clinic-centric treatment, while the patient is at the clinic, to patient-centric, when the patient is discharged but continues the treatment at home or elsewhere, all while connected to each other at all times. The condition of the patient could then be monitored by healthcare providers from the clinic without having to visit the patient in person. Having an intelligent assisted living environment for home-based healthcare as presented by \cite{ThirtySeven} would also enable caregivers to tend to specific needs at any time. Given the criticality of the IoT-based medical devices in such a scenario, it is important that these devices work correctly at all times and provide correct insights. To ensure that these devices work correctly, they would have to undergo periodic preventive maintenance. The authors of \cite{ThirtyThree} investigated this and proposed autonomous integrity monitoring architecture that would provide complete visibility into the medical devices as well as predict potential failures before they actually happen. The architecture utilizes big data technologies to autonomously diagnose the components for wear and tear. \\

The number of applications and how ML and IoT in healthcare are being combined is growing every day. Looking at all the applications discussed in this section, we get a good overview and an answer to RQ1.

\subsection{Collaborative Machine Learning \& IoMT}
While we explored the different ways in which ML and IoT in healthcare are being combined today in section 3.1, we observed that all the works mentioned some form of collaborative computing. Both sound very similar but have a few differences. To be precise, 67\% of the primary studies discussed or made use of distributed machine learning, and the remaining 33\% proposed solutions involving distributed data processing and employing machine learning at a later stage. With the majority of the primary studies directly addressing distributed machine learning, it is evident that distributed machine learning can be applied to IoT healthcare data, answering RQ2.

\begin{figure}[H]
    \centering
    \includegraphics[width=8.7cm]{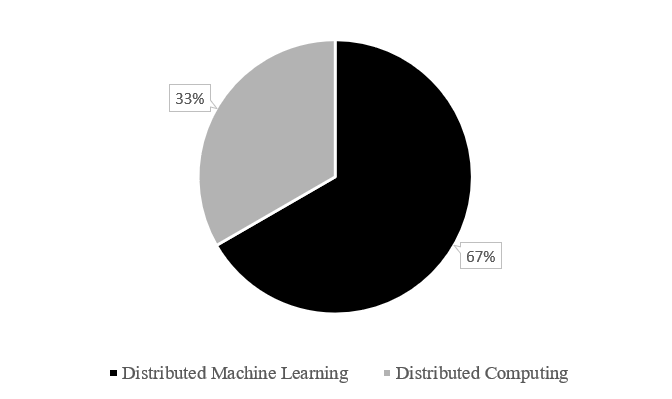}
    \caption{Distributed ML vs Distributed Processing}
    \label{fig:dml_dc}
\end{figure}

\subsection{Distributed Machine Learning Algorithms}
Having explored the various applications of ML in healthcare IoT in section 3.1 and confirmed the possibility of applying distributed machine learning techniques to the same in section 3.2, this section evaluates the distributed machine learning algorithms and distributed computing frameworks that have been proposed in the primary studies. The best performing algorithms, in terms of prediction accuracy, are then identified for each healthcare scenario - live health monitoring systems, and medical diagnostic systems.

\subsubsection{\textbf{Live Health Monitoring Systems}}
The growing number of people suffering from chronic illnesses has led to the development and innovation of several live health monitoring systems. IoT-based live health monitoring systems are essential for such people to keep their health in check, and this could be a life or death scenario. Considering how critical such systems and their applications are, they need to produce accurate predictions in real-time. For benchmarking the standard model for IoT healthcare systems proposed by \cite{Seven}, a few ML algorithms were trained in the cloud on blood oxygen level data to monitor people's health and identify underlying conditions. The predictions produced were reasonably accurate compared to the actual results of patients undergoing surgery in a clinical setting. 

The evaluation of ML algorithms, namely K-NN, Support Vector Machine, Decision Trees, Random Forest, and MLP, on publicly available datasets for various diseases by \cite{Eight} offered insights into their performances. Random Forest outperformed the rest in most scenarios, achieving an accuracy of 96.42\% for breast cancer, 97.26\% in dermatology, and 81.16\% for diabetes. The accuracy could be further improved by working with larger training datasets. To test the performance of distributed processing framework for the smartwatch app developed by \cite{Sixteen}, the authors tested four algorithms - Random Forest, Gradient-boosted Tree, Logistic Regression, and Support Vector Machine and compared their F1 scores. Gradient-boosted tree performed the best with an F1 score of 0.983 but took the most time to train, with Random Forest coming second at 0.977 and taking the least time to train.

The authors of \cite{Thirteen} discussed the applicability of several ML algorithms on different kinds of healthcare datasets and concluded that none of them are tailored for IoT applications. This led to the development of a custom shallow Artificial Neural Network for ML on-chip for low-end devices. The ANN achieved an accuracy of 95\%, which was comparable to the performance of a Convolutional Neural Network trained on the same ECG data on an edge node.

Study \cite{ThirtyTwo} put the HiCH architecture, built by \cite{TwentyEight}, to test by training a CNN classifier on ECG data from MIT's Arrhythmia dataset first to set a baseline. Then it was retrained on incoming ECG signals from the patient to be monitored to tailor the accuracy to the patient. With this method, the authors managed to achieve an accuracy of 96\%. While tailoring algorithms to the end-users yields higher accuracy, this approach is not feasible for time-critical applications. Similarly, the authors of \cite{Six} built upon the three-layer distributed IoT architecture proposed by \cite{Eighteen} and came up with a system to track the location and vitals of soldiers on the battlefield in real-time. K-Means Clustering was the algorithm of choice which classified different activities. The model produced results well enough to align with real-world scenarios, but accuracy numbers were not reported. \\

Looking at the overall findings from all the studies on live health monitoring systems, the majority of the studies concluded that Random Forest performed the best in terms of overall accuracy.

\subsubsection{\textbf{Medical Diagnostic Systems}}
Diagnosing health and vitals is as important as monitoring them. The authors of the study \cite{Nine} designed an architecture for smart healthcare focused on diagnosing heart-related diseases. Four different classification algorithms - Naive Bayes, k-Nearest Neighbor, Decision Tree, and Support Vector Machine,  were tested. Decision Tree came out at the top-yielding an accuracy of 92.8\%, and also yielded the highest values for specificity, sensitivity, and F-measure. 

The cutting-edge cloud-based teleophthalmology framework built by \cite{Eleven} was designed to utilize AMD ResNet, which is a modified ResNet Convolutional Neural Network and a long–short term memory-based neural network in one. Upon testing, the model achieved an accuracy of 79.4\%, which is significantly higher than the accuracy of an expert ophthalmologist. The HealthFog framework by \cite{ThirtyFive}, built for diagnosing heart health and disease severity, used an ensemble deep neural network for analysis and binary classification. The neural network achieved an accuracy of 89\%, courtesy of how ensemble learning works. 

The FedHealth framework built by \cite{FiftyOne} took the personalized healthcare model approach similar to the one taken by \cite{TwentyEight} and brought it to wearable devices, utilizing a CNN and optimizing with Stochastic Gradient Descent. The authors tested the framework's performance through fine-tuning and a transfer with maximum mean discrepancy - the first being a tuning approach and the second transfer learning. Transfer learning outperformed by 4\% in terms of average accuracy, with the authors claiming FedHealth produced higher accuracy for wearable activity recognition when compared to traditional methods \cite{FiftyOne}. \\

Looking at the overall findings from all the studies on medical diagnostic systems, the majority of the studies proposed their modified versions of Neural Networks which, as a result, helped attain the best accuracy numbers for each. No existing state-of-the-art machine learning algorithm stood out.

%% file: discussion.tex
\section{Discussion} 
Collaborative (or distributed) machine learning algorithms were the focus of this review. The research goal was the identification of algorithms that performed the best in the IoT healthcare scenarios considered - live health monitoring systems and medical diagnostic systems. Completing the review gave us the answers we were looking for, but the results are not definitive i.e., Random Forest may not always achieve the highest accuracy when deployed in live monitoring systems. A few studies used other algorithms for their case studies and had different conclusions. Some even built their neural networks to achieve higher accuracy, the majority taking this approach for diagnostic systems. 

Overall, no known algorithm performs consistently across all scenarios in healthcare as there are numerous variables involved. However, Random Forest could serve as a good starting point for distributed machine learning applications in IoMT. With the growing number of machine learning-driven IoT healthcare applications, more technical contributions to this subject area are expected with more conclusive results.

%% file: conclusion.tex
\section{Conclusion}
There is an increasing demand for IoT systems and devices due to their potential in healthcare. IoMT technology has been aiding people with chronic illnesses in various ways, improving the quality of their lives. Making these systems smarter by introducing machine learning has great potential, with the focus shifting to obtaining faster real-time predictions over the years. Researchers have been experimenting with various distributed computing frameworks to achieve the highest possible prediction accuracy while being efficient in terms of resources at the same time. 

This systematic literature review started with a broad scope of exploring how the healthcare industry had been combining machine learning and IoT. It then narrowed down to determining whether distributed machine learning could be applied to IoT healthcare data or not. Then narrowed further down to identifying and comparing popular distributed machine learning algorithms and frameworks being used in various state-of-the-art IoT healthcare systems.

Upon analyzing the primary studies, they were classified into three categories (scenarios) - monitoring health and vitals, diagnosing health conditions, and other innovative yet noteworthy systems being used in the healthcare industry today. Each of the primary studies touched upon distributed computing and distributed machine learning, with the latter being used by the majority. The studies discussing distributed computing were similar and utilized machine learning using the distributed frameworks proposed by the authors of the respective studies. Then the machine learning algorithms tested in the studies and their prediction accuracy were evaluated, giving us Random Forest as the algorithm that achieved the highest accuracy in a few cases. However, most studies proposed their own variant of a neural network tailored to their specific case studies and frameworks. This subject area is still fairly new with many contributions being made. It will take a few more years until we see state-of-the-art algorithms for IoMT scenarios.

%% file: main.bbl

\begin{thebibliography}{24}

%%% the Web address.

\ifx \showCODEN    \undefined \def \showCODEN     #1{\unskip}     \fi
\ifx \showDOI      \undefined \def \showDOI       #1{#1}\fi
\ifx \showISBNx    \undefined \def \showISBNx     #1{\unskip}     \fi
\ifx \showISBNxiii \undefined \def \showISBNxiii  #1{\unskip}     \fi
\ifx \showISSN     \undefined \def \showISSN      #1{\unskip}     \fi
\ifx \showLCCN     \undefined \def \showLCCN      #1{\unskip}     \fi
\ifx \shownote     \undefined \def \shownote      #1{#1}          \fi
\ifx \showarticletitle \undefined \def \showarticletitle #1{#1}   \fi
\ifx \showURL      \undefined \def \showURL       {\relax}        \fi
% invisible to TeX
\providecommand\bibfield[2]{#2}
\providecommand\bibinfo[2]{#2}
\providecommand\natexlab[1]{#1}
\providecommand\showeprint[2][]{arXiv:#2}

\bibitem[\protect\citeauthoryear{Azimi, Anzanpour, Rahmani, Pahikkala,
  Levorato, Liljeberg, and Dutt}{Azimi et~al\mbox{.}}{2017}]%
        {TwentyEight}
\bibfield{author}{\bibinfo{person}{Iman Azimi}, \bibinfo{person}{Arman
  Anzanpour}, \bibinfo{person}{Amir~M. Rahmani}, \bibinfo{person}{Tapio
  Pahikkala}, \bibinfo{person}{Marco Levorato}, \bibinfo{person}{Pasi
  Liljeberg}, {and} \bibinfo{person}{Nikil Dutt}.}
  \bibinfo{year}{2017}\natexlab{}.
\newblock \showarticletitle{HiCH: Hierarchical Fog-Assisted Computing
  Architecture for Healthcare IoT}.
\newblock  (\bibinfo{year}{2017}).
\newblock


\bibitem[\protect\citeauthoryear{Azimi, Takalo-Mattila, Anzanpour, Rahmani,
  Soininen, and Liljeberg}{Azimi et~al\mbox{.}}{2018}]%
        {ThirtyTwo}
\bibfield{author}{\bibinfo{person}{Iman Azimi}, \bibinfo{person}{Janne
  Takalo-Mattila}, \bibinfo{person}{Arman Anzanpour}, \bibinfo{person}{Amir~M.
  Rahmani}, \bibinfo{person}{Juha-Pekka Soininen}, {and} \bibinfo{person}{Pasi
  Liljeberg}.} \bibinfo{year}{2018}\natexlab{}.
\newblock \showarticletitle{Empowering Healthcare IoT Systems with Hierarchical
  Edge-based Deep Learning}.
\newblock  (\bibinfo{year}{2018}).
\newblock


\bibitem[\protect\citeauthoryear{Baker, Xiang, and Atkinson}{Baker
  et~al\mbox{.}}{2017}]%
        {Seven}
\bibfield{author}{\bibinfo{person}{Stephanie~B. Baker}, \bibinfo{person}{Wei
  Xiang}, {and} \bibinfo{person}{Ian Atkinson}.}
  \bibinfo{year}{2017}\natexlab{}.
\newblock \showarticletitle{Internet of Things for Smart Healthcare:
  Technologies, Challenges, and Opportunities}.
\newblock  (\bibinfo{year}{2017}).
\newblock


\bibitem[\protect\citeauthoryear{Bandopadhaya, Dey, and Suhag}{Bandopadhaya
  et~al\mbox{.}}{2020}]%
        {Eighteen}
\bibfield{author}{\bibinfo{person}{Shuvabrata Bandopadhaya},
  \bibinfo{person}{Rajiv Dey}, {and} \bibinfo{person}{Ashok Suhag}.}
  \bibinfo{year}{2020}\natexlab{}.
\newblock \showarticletitle{Integrated healthcare monitoring solutions for
  soldier using the internet of things with distributed computing}.
\newblock  (\bibinfo{year}{2020}).
\newblock


\bibitem[\protect\citeauthoryear{Chen, Wang, Yu, Gao, and Qin}{Chen
  et~al\mbox{.}}{2019}]%
        {FiftyOne}
\bibfield{author}{\bibinfo{person}{Yiqiang Chen}, \bibinfo{person}{Jindong
  Wang}, \bibinfo{person}{Chaohui Yu}, \bibinfo{person}{Wen Gao}, {and}
  \bibinfo{person}{Xin Qin}.} \bibinfo{year}{2019}\natexlab{}.
\newblock \showarticletitle{FedHealth: A Federated Transfer Learning Framework
  for Wearable Healthcare}.
\newblock  (\bibinfo{year}{2019}).
\newblock


\bibitem[\protect\citeauthoryear{Das, Rad, Choo, Nouhi, Lish, and Martel}{Das
  et~al\mbox{.}}{2019}]%
        {Eleven}
\bibfield{author}{\bibinfo{person}{Arun Das}, \bibinfo{person}{Paul Rad},
  \bibinfo{person}{Kim-Kwang~Raymond Choo}, \bibinfo{person}{Babak Nouhi},
  \bibinfo{person}{Jonathan Lish}, {and} \bibinfo{person}{James Martel}.}
  \bibinfo{year}{2019}\natexlab{}.
\newblock \showarticletitle{Distributed machine learning cloud
  teleophthalmology IoT for predicting AMD disease progression}.
\newblock  (\bibinfo{year}{2019}).
\newblock


\bibitem[\protect\citeauthoryear{{Dyba}, {Dingsoyr}, and {Hanssen}}{{Dyba}
  et~al\mbox{.}}{2007}]%
        {ApplyingSystematicReviews}
\bibfield{author}{\bibinfo{person}{T. {Dyba}}, \bibinfo{person}{T. {Dingsoyr}},
  {and} \bibinfo{person}{G.~K. {Hanssen}}.} \bibinfo{year}{2007}\natexlab{}.
\newblock \showarticletitle{Applying Systematic Reviews to Diverse Study Types:
  An Experience Report}. In \bibinfo{booktitle}{\emph{First International
  Symposium on Empirical Software Engineering and Measurement (ESEM 2007)}}.
  \bibinfo{pages}{225--234}.
\newblock


\bibitem[\protect\citeauthoryear{Ed-daoudy and Maalmi}{Ed-daoudy and
  Maalmi}{2019}]%
        {FortyOne}
\bibfield{author}{\bibinfo{person}{Abderrahmane Ed-daoudy} {and}
  \bibinfo{person}{Khalil Maalmi}.} \bibinfo{year}{2019}\natexlab{}.
\newblock \showarticletitle{A new Internet of Things architecture for real-time
  prediction of various diseases using machine learning on big data
  environment}.
\newblock  (\bibinfo{year}{2019}).
\newblock


\bibitem[\protect\citeauthoryear{Farahani, Barzegari, and Aliee}{Farahani
  et~al\mbox{.}}{2019}]%
        {Thirteen}
\bibfield{author}{\bibinfo{person}{Bahar Farahani}, \bibinfo{person}{Mojtaba
  Barzegari}, {and} \bibinfo{person}{Fereidoon~Shams Aliee}.}
  \bibinfo{year}{2019}\natexlab{}.
\newblock \showarticletitle{Towards Collaborative Machine Learning Driven
  Healthcare Internet of Things}.
\newblock  (\bibinfo{year}{2019}).
\newblock


\bibitem[\protect\citeauthoryear{Farahani, Firouzi, Chang, Badaroglu, Constant,
  and Mankodiya}{Farahani et~al\mbox{.}}{2018}]%
        {TwentyFive}
\bibfield{author}{\bibinfo{person}{Bahar Farahani}, \bibinfo{person}{Farshad
  Firouzi}, \bibinfo{person}{Victor Chang}, \bibinfo{person}{Mustafa
  Badaroglu}, \bibinfo{person}{Nicholas Constant}, {and} \bibinfo{person}{Kunal
  Mankodiya}.} \bibinfo{year}{2018}\natexlab{}.
\newblock \showarticletitle{Towards fog-driven IoT eHealth: Promises and
  challenges of IoT in medicine and healthcare}.
\newblock  (\bibinfo{year}{2018}).
\newblock


\bibitem[\protect\citeauthoryear{Fonseca, Mendes, Lopes, Romão, and
  Parreira}{Fonseca et~al\mbox{.}}{2017}]%
        {ThirtySeven}
\bibfield{author}{\bibinfo{person}{Césa Fonseca}, \bibinfo{person}{David
  Mendes}, \bibinfo{person}{Manuel Lopes}, \bibinfo{person}{Artur Romão},
  {and} \bibinfo{person}{Pedro Parreira}.} \bibinfo{year}{2017}\natexlab{}.
\newblock \showarticletitle{Deep Learning and IoT to Assist Multimorbidity Home
  Based Healthcare}.
\newblock  (\bibinfo{year}{2017}).
\newblock


\bibitem[\protect\citeauthoryear{Fozoonmayeh, Le, Wittfoth, Geng, Ha, Wang,
  Vasilenko, Ahn, and kyung Woodbridge}{Fozoonmayeh et~al\mbox{.}}{2020}]%
        {Sixteen}
\bibfield{author}{\bibinfo{person}{Donya Fozoonmayeh}, \bibinfo{person}{Hai~Vu
  Le}, \bibinfo{person}{Ekaterina Wittfoth}, \bibinfo{person}{Chong Geng},
  \bibinfo{person}{Natalie Ha}, \bibinfo{person}{Jingjue Wang},
  \bibinfo{person}{Maria Vasilenko}, \bibinfo{person}{Yewon Ahn}, {and}
  \bibinfo{person}{Diane~Myung kyung Woodbridge}.}
  \bibinfo{year}{2020}\natexlab{}.
\newblock \showarticletitle{A Scalable Smartwatch-Based Medication Intake
  Detection System Using Distributed Machine Learning}.
\newblock  (\bibinfo{year}{2020}).
\newblock


\bibitem[\protect\citeauthoryear{Gondalia, Dixit, Parashar, Raghava, and
  Sengupta}{Gondalia et~al\mbox{.}}{2018}]%
        {Six}
\bibfield{author}{\bibinfo{person}{Aashay Gondalia}, \bibinfo{person}{Dhruv
  Dixit}, \bibinfo{person}{Shubham Parashar}, \bibinfo{person}{Vijayanand
  Raghava}, {and} \bibinfo{person}{Animesh Sengupta}.}
  \bibinfo{year}{2018}\natexlab{}.
\newblock \showarticletitle{IoT-based Healthcare Monitoring System for War
  Soldiers using Machine Learning}.
\newblock  (\bibinfo{year}{2018}).
\newblock


\bibitem[\protect\citeauthoryear{Greco, Percannella, Ritrovato, Tortorella, and
  Vento}{Greco et~al\mbox{.}}{2020}]%
        {ThirtySix}
\bibfield{author}{\bibinfo{person}{Luca Greco}, \bibinfo{person}{Gennaro
  Percannella}, \bibinfo{person}{Pierluigi Ritrovato},
  \bibinfo{person}{Francesco Tortorella}, {and} \bibinfo{person}{Mario Vento}.}
  \bibinfo{year}{2020}\natexlab{}.
\newblock \showarticletitle{Trends in IoT based solutions for health care:
  Moving AI to the edge}.
\newblock  (\bibinfo{year}{2020}).
\newblock


\bibitem[\protect\citeauthoryear{Halevi, Moed, and Bar-Ilan}{Halevi
  et~al\mbox{.}}{2017}]%
        {GoogleScholar}
\bibfield{author}{\bibinfo{person}{Gali Halevi}, \bibinfo{person}{Henk Moed},
  {and} \bibinfo{person}{Judit Bar-Ilan}.} \bibinfo{year}{2017}\natexlab{}.
\newblock \showarticletitle{Suitability of Google Scholar as a source of
  scientific information and as a source of data for scientific
  evaluation—Review of the Literature}.
\newblock  (\bibinfo{year}{2017}).
\newblock


\bibitem[\protect\citeauthoryear{Kaur, Kumar, and Kumar}{Kaur
  et~al\mbox{.}}{2019}]%
        {Eight}
\bibfield{author}{\bibinfo{person}{Pavleen Kaur}, \bibinfo{person}{Ravinder
  Kumar}, {and} \bibinfo{person}{Munish Kumar}.}
  \bibinfo{year}{2019}\natexlab{}.
\newblock \showarticletitle{A healthcare monitoring system using random forest
  and internet of things (IoT)}.
\newblock  (\bibinfo{year}{2019}).
\newblock


\bibitem[\protect\citeauthoryear{Kitchenham and Brereton}{Kitchenham and
  Brereton}{2013}]%
        {SysReviewProc}
\bibfield{author}{\bibinfo{person}{Barbara Kitchenham} {and}
  \bibinfo{person}{Pearl Brereton}.} \bibinfo{year}{2013}\natexlab{}.
\newblock \showarticletitle{A Systematic Review of Systematic Review Process
  Research in Software Engineering}.
\newblock  \bibinfo{volume}{55}, \bibinfo{number}{12} (\bibinfo{year}{2013}),
  \bibinfo{pages}{2049–2075}.
\newblock


\bibitem[\protect\citeauthoryear{Maktoubian and Ansari}{Maktoubian and
  Ansari}{2019}]%
        {ThirtyThree}
\bibfield{author}{\bibinfo{person}{Jamal Maktoubian} {and}
  \bibinfo{person}{Keyvan Ansari}.} \bibinfo{year}{2019}\natexlab{}.
\newblock \showarticletitle{An IoT architecture for preventive maintenance of
  medical devices in healthcare organizations}.
\newblock  (\bibinfo{year}{2019}).
\newblock


\bibitem[\protect\citeauthoryear{Palanisamy and Thirunavukarasu}{Palanisamy and
  Thirunavukarasu}{2019}]%
        {FortyFour}
\bibfield{author}{\bibinfo{person}{Venketesh Palanisamy} {and}
  \bibinfo{person}{Ramkumar Thirunavukarasu}.} \bibinfo{year}{2019}\natexlab{}.
\newblock \showarticletitle{Implications of big data analytics in developing
  healthcare frameworks – A review}.
\newblock  (\bibinfo{year}{2019}).
\newblock


\bibitem[\protect\citeauthoryear{Rathore, Ahmad, Paul, Wan, and Zhang}{Rathore
  et~al\mbox{.}}{2016}]%
        {Fifty}
\bibfield{author}{\bibinfo{person}{M.~Mazhar Rathore}, \bibinfo{person}{Awais
  Ahmad}, \bibinfo{person}{Anand Paul}, \bibinfo{person}{Jiafu Wan}, {and}
  \bibinfo{person}{Daqiang Zhang}.} \bibinfo{year}{2016}\natexlab{}.
\newblock \showarticletitle{Real-time Medical Emergency Response System:
  Exploiting IoT and Big Data for Public Health}.
\newblock  (\bibinfo{year}{2016}).
\newblock


\bibitem[\protect\citeauthoryear{Tuli, Basumatary, Gill, Kahani, Arya, Wander,
  and Buyya}{Tuli et~al\mbox{.}}{2020}]%
        {ThirtyFive}
\bibfield{author}{\bibinfo{person}{Shreshth Tuli}, \bibinfo{person}{Nipam
  Basumatary}, \bibinfo{person}{Sukhpal~Singh Gill}, \bibinfo{person}{Mohsen
  Kahani}, \bibinfo{person}{Rajesh~Chand Arya}, \bibinfo{person}{Gurpreet~Singh
  Wander}, {and} \bibinfo{person}{Rajkumar Buyya}.}
  \bibinfo{year}{2020}\natexlab{}.
\newblock \showarticletitle{HealthFog: An ensemble deep learning based Smart
  Healthcare System for Automatic Diagnosis of Heart Diseases in integrated IoT
  and fog computing environments}.
\newblock  (\bibinfo{year}{2020}).
\newblock


\bibitem[\protect\citeauthoryear{Verma and Sood}{Verma and Sood}{2017}]%
        {Nine}
\bibfield{author}{\bibinfo{person}{Prabal Verma} {and}
  \bibinfo{person}{Sandeep~K. Sood}.} \bibinfo{year}{2017}\natexlab{}.
\newblock \showarticletitle{Cloud-centric IoT based disease diagnosis
  healthcare framework}.
\newblock  (\bibinfo{year}{2017}).
\newblock


\bibitem[\protect\citeauthoryear{Yuan, Ge, and Xing}{Yuan
  et~al\mbox{.}}{2020}]%
        {Twenty}
\bibfield{author}{\bibinfo{person}{Binhang Yuan}, \bibinfo{person}{Song Ge},
  {and} \bibinfo{person}{Wenhui Xing}.} \bibinfo{year}{2020}\natexlab{}.
\newblock \showarticletitle{A Federated Learning Framework for Healthcare IoT
  devices}.
\newblock  (\bibinfo{year}{2020}).
\newblock


\bibitem[\protect\citeauthoryear{Zheng, Sun, Mukkamala, Vatrapu, and
  Ordieres-Meré}{Zheng et~al\mbox{.}}{2019}]%
        {TwentyNine}
\bibfield{author}{\bibinfo{person}{Xiaochen Zheng}, \bibinfo{person}{Shengjing
  Sun}, \bibinfo{person}{Raghava~Rao Mukkamala}, \bibinfo{person}{Ravi
  Vatrapu}, {and} \bibinfo{person}{Joaquín Ordieres-Meré}.}
  \bibinfo{year}{2019}\natexlab{}.
\newblock \showarticletitle{Accelerating Health Data Sharing: A Solution Based
  on the Internet of Things and Distributed Ledger Technologies}.
\newblock  (\bibinfo{year}{2019}).
\newblock


\end{thebibliography}
